\pdfoutput=1

\documentclass[11pt]{article}

\usepackage{acl}

\usepackage{times}
\usepackage{latexsym}
\usepackage{amsmath}
\usepackage{amsfonts}
\usepackage{amssymb}
\usepackage{tabularx}
\usepackage{booktabs}
\usepackage{multirow}

\usepackage[T1]{fontenc}

\usepackage[utf8]{inputenc}

\usepackage{microtype}

\usepackage{inconsolata}
\usepackage{graphicx}
\usepackage{cleveref}
\newcommand{\Scref}[1]{\S\ref{#1}}

\title{Towards Reliable and Empathetic Depression-Diagnosis-Oriented Chats}

\author{Kunyao Lan$^{1,2,3}$, Cong Ming$^{1,2,3}$, Binwei Yao$^4$, Lu Chen$^{1,2,3}$, Mengyue Wu$^{1,2,3}$\\
$^{1}$X-LANCE Lab, Dept. of Computer Science and Engineering\\$^{2}$MoE Key Lab of Artificial Intelligence, AI Institute\\$^{3}$Shanghai Jiao Tong University, China\\
$^{4}$University of Wisconsin - Madison, U.S.\\
\texttt{\{lankunyao, chenlusz,mengyuewu\}@sjtu.edu.cn,}\\
\texttt{mcheather19@outlook.com} \\
\texttt{binwei.yao@wisc.edu}
}

\begin{document}
\maketitle
\begin{abstract}
Chatbots can serve as a viable tool for preliminary depression diagnosis via interactive conversations with potential patients. Nevertheless, the blend of task-oriented and chit-chat in diagnosis-related dialogues necessitates professional expertise and empathy. Such unique requirements challenge traditional dialogue frameworks geared towards single optimization goals. To address this, we propose an innovative ontology definition and generation framework tailored explicitly for depression diagnosis dialogues, combining the reliability of task-oriented conversations with the appeal of empathy-related chit-chat. We further apply the framework to D$^4$, the only existing public dialogue dataset on depression diagnosis-oriented chats. Exhaustive experimental results indicate significant improvements in task completion and emotional support generation in depression diagnosis, fostering a more comprehensive approach to task-oriented chat dialogue system development and its applications in digital mental health.
\end{abstract}

\section{Introduction}

Dialogue agents are promising diagnostic tools for large-scale depression screening~\cite{pacheco2021smart}.
They can potentially reduce the concealment of sensitive information~\cite{schuetzler2018influence} and overcome emotional barriers in face-to-face conversations~\cite{hart2017virtual}.
During diagnosing depression, doctors ask patients about symptoms while providing emotional support to elicit sufficient self-expression. 
Therefore, unlike traditional widely-used dialogue systems, depression-diagnosis-oriented chats require appropriate emotional support~\cite{liu-etal-2021-towards} during diagnosis conversations. This approach combines task-oriented dialogue and chit-chat characteristics, named Task-Oriented Chat (TOC)~\cite{yao-etal-2022-d4}.
\begin{figure}[t!]
   \centering
   \noindent\includegraphics[width=\columnwidth]{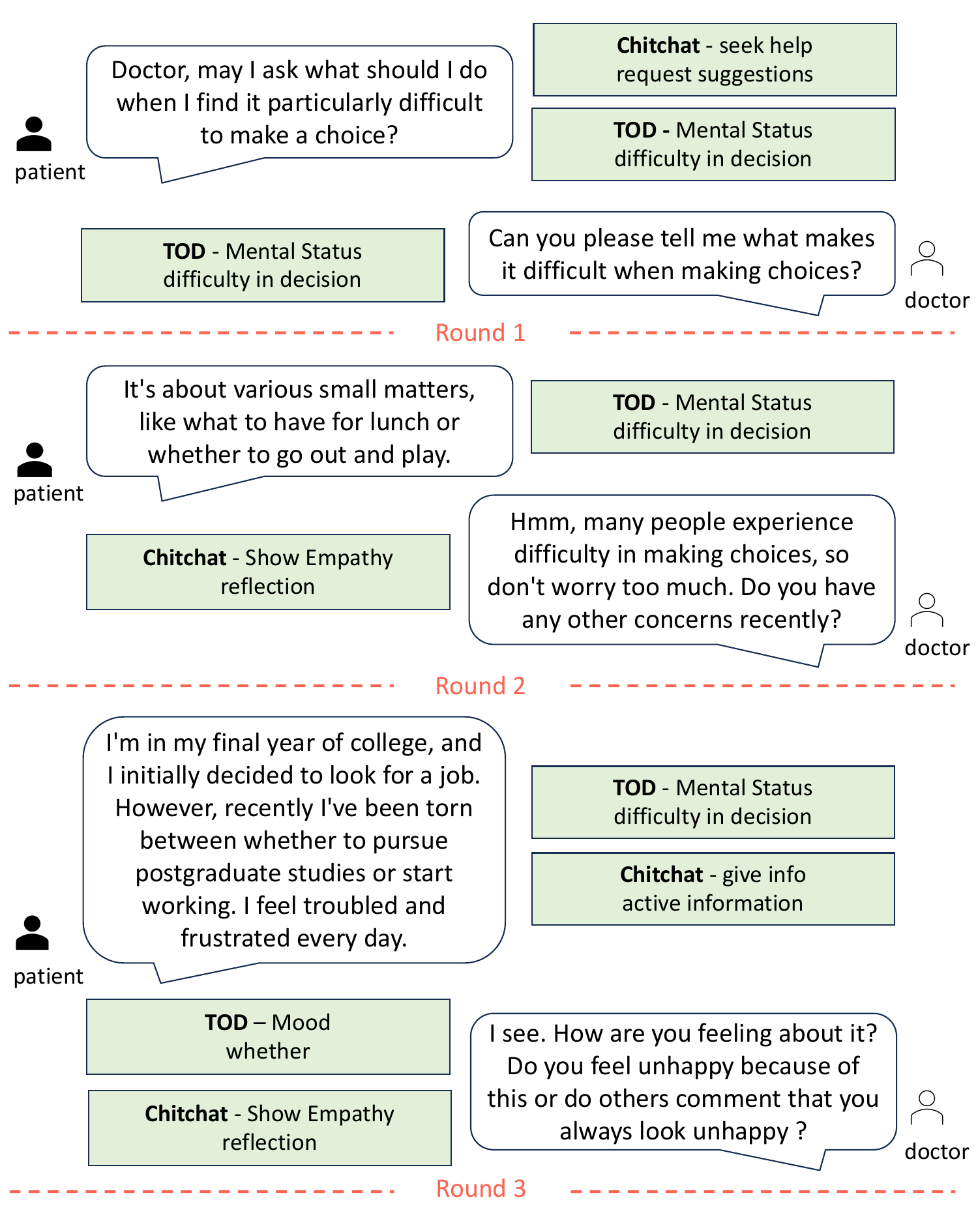}
   \caption{Task-Oriented Chat in Depression Diagnosis, complex intents with flexible intent transition, simultaneously enveloping TOD and Chit-chat in single turn.}
   \label{fig: annotation}
\end{figure}

However, current dialogue systems still struggle to meet the high demands for \textit{accuracy} and \textit{empathy} in depression-diagnosis dialogues.
First, compared to traditional task-oriented dialogue systems, depression diagnostic dialogue demands high accuracy in task completion for diagnosis purposes. However, the current systems often omit core symptom acquirement and jump to an inaccurate diagnostic conclusion, further lacking an evaluation mechanism to assess the accuracy of diagnostic tasks, which is crucial for the practical application of diagnosis chatbots.

Second, current dialogue systems have difficulty in delivering \textit{rich} and \textit{appropriate empathetic responses}. This limitation restricts systems' ability to engage users effectively and hampers the construction of therapeutic alliance. 
Dialogue systems face these challenges because they are traditionally optimized for a single objective. Task-oriented systems focused on relatively short dialogues involving specific tasks~\cite{budzianowski2018multiwoz} like ticket booking or navigation, while chit-chat systems aimed to mimic human-like conversations for long-term engagement and meeting emotional needs~\cite{huang2020challenges}. This divergence in objectives makes it difficult for previous systems to achieve multiple composite goals simultaneously.

To address this, we propose a novel depression diagnosis ontology framework named \textbf{SEO}, which comprises two distinct components: \textbf{S}ymptom-related (serves TOD) and \textbf{E}mpathy-related (tailored for chit-chat) \textbf{O}ntology, as shown in \Cref{fig: annotation}. The example has been translated from Chinese into English for better understanding. The ontology comprises symptom-related ontology based on depression diagnosis criteria DSM-5~\cite{APA2013} and empathy-related ontology by Helping Skills Theory~\cite{huang2020challenges, hill2020helping}. Although there are several ontology framework related to mental health, they either focus only on symptoms~\cite{song-etal-2023-simple}, ignoring the construction of therapeutic alliance and hampering the patient expressing their difficulties or detailed symptoms, or focus mainly on emotional support~\cite{huang2020challenges,li-etal-2023-understanding}, lacking the ability of diagnosis.

SEO integrates symptom and empathy strategies but \textbf{\textit{goes beyond simple combination}}: the flexible and dynamic transition between the two aspects is essential as chit-chat interactions form an inherent part of the diagnostic process, and without a properly defined structure bridging the two parts, the full scope of real-world interactive nuances in depression diagnosis cannot be adequately understood or utilized. This dynamic integration not only enriches the diagnostic dialogue but also ensures a more holistic and authentic engagement with patients, crucial for effective diagnosis and treatment.
Extensive experiments from generation tasks to risk classification along with human evaluation results showed that our ontology helps build a robust and empathetic depression-diagnosis-oriented dialogue system.
The key contribution of this paper is as follows:
\begin{itemize}
    \setlength{\itemsep}{0pt}
    \setlength{\parsep}{0pt}
    \setlength{\parskip}{0pt}
    \item Our proposed framework SEO integrates symptom and empathy ontologies in depression diagnosis, essential for comprehensively understanding and utilizing real-world interactive nuances vital for effective patient engagement and accurate diagnosis. 
    \item For the first time, TOC is formally defined in a real-world dialogue scenario by jointing TOD and chit-chat components in depression-diagnosis-chat and analyzing the cross-domain intent combination and flexible transitions in this complex dialogue scene. 
    \item Experimental validation on the annotated datasets demonstrates that the new ontology enables pre-training sequence-to-sequence models to effectively achieve multiple objectives in TOC.
\end{itemize}

\begin{figure}[t]
   \centering
   \noindent\includegraphics[width=0.8\columnwidth]{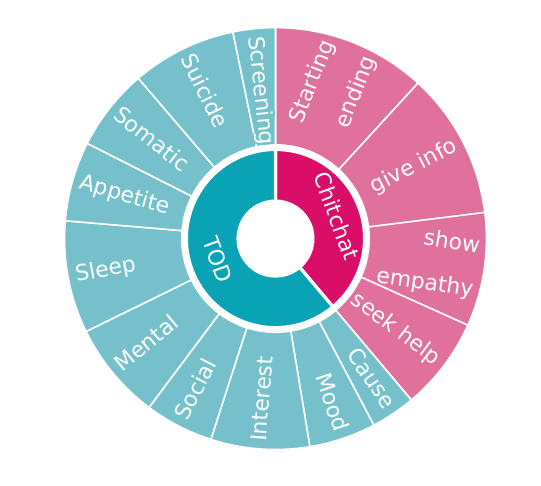}
   \caption{SEO Intent Distribution of D$^4$. The length along the circumference serves as an indicator of the proportional representation of the intent. Note that a 3-level intent hierarchy is maintained, the complete intent list is provided in Appendix \Cref{tab:ontology}.}
   \label{fig:sunburst}
\end{figure}

\section{SEO Ontology Framework}

This section presents our methodology of Task-Oriented Chat ontology definition of depression diagnosis and its annotations on Chinese dataset D$^4$, the sole publicly available depression diagnosis dialogue dataset. In order to overcome the prolonged challenges in depression diagnostic dialogue systems and leverage the strengths of both Task-Oriented Dialogue (TOD) and Chit-Chat, we propose a novel depression diagnosis ontology framework named \textbf{SEO}: \textbf{S}ymptom-related (serves TOD) and \textbf{E}mpathy-related (tailored for chit-chat) \textbf{O}ntology.

\subsection{Ontology Definition}
\label{subsec:ontology definition}

The SEO comprises two distinct components: symptom-related TOD ontology and empathy-related chit-chat ontology. A detailed representation of SEO is presented in appendix \Cref{tab:ontology}.

These components serve different purposes, with the symptom-related TOD ontology emphasizing the need for accurate diagnosis and the empathy-related chit-chat ontology focusing on providing practical support. Consequently, the foundation of the symptom-related TOD ontology lies in the DSM-5~\cite{APA2013},  which represents one of the most widely adopted diagnostic criteria for depression in clinical practice. On the other hand, the empathy-related chit-chat ontology draws its inspiration from the Helping Skills Theory~\cite{hill2020helping}, which is renowned for its advanced strategies in offering emotional support and has been extensively utilized in previous works~\cite{liu2021towards}.

\paragraph{Symptom-related TOD Ontology} Currently, the widely utilized definitions of symptom-related ontology primarily concentrate on core depressive symptoms while entrusting the assessment of symptom severity to the patients. While these ontology definitions enhance efficiency and reduce costs, they also introduce the risk of inaccurate severity ratings, imprecise questioning, and ambiguous expressions~\cite{lee2010current}. Hence, it becomes crucial to refine the criteria associated with each core symptom to attain a more precise evaluation of the severity of fundamental symptoms. 

Symptom-related TOD ontology encompasses two levels: \textbf{\textit{core symptoms}} and \textbf{\textit{fine-grained symptoms}} associated with them. Initially, we delineate ten core symptoms of depression based on previous research~\cite{yao-etal-2022-d4} and DSM-5 guidelines. These core symptoms include \textit{cause, mood, interest, social function, mental status, sleep, appetite, somatic symptoms, suicide, and screening}. Additionally, in order to indicate that the respondent clearly describes how he/she has been feeling these days, the core symptoms need to be subdivided into their respective components~\cite{https://doi.org/10.1002/jclp.10207}, which also improves the performance of discriminating the depression severity of patients~\cite{10.3389/fpsyt.2018.00450}. Therefore, we identify one to six relevant fine-grained symptoms for each core symptom as shown in \Cref{fig:sunburst}
, resulting in a comprehensive symptom-related ontology. 

\paragraph{Empathy-related Chit-chat Ontology}
In depression diagnosis dialogues, building trust is as crucial as gathering symptoms, making empathy strategies crucial in chit-chat ontology. Previous research treated empathy as a separate dialogue strategy from symptom-related ones, leading to lower generation quality when psychiatrists use various empathy strategies. To address this, we've broadened the empathy concept into a series of strategies, enhancing empathetic response generation. We also incorporate emotion recognition to help the model select appropriate empathy strategies.

Our definition of empathy-related chit-chat ontology, based on prior studies, includes several empathy dialogue strategies. These strategies are divided into two main parts: \textbf{\textit{empathy}} strategies (reflection, self-disclosure, affirmation, providing suggestions) focus on building trust with the patient, and \textbf{\textit{transition}} strategies (starting/ending, requiring personal information, restatement) manage dialogue flow. This ontology equips the chatbot with a range of empathy strategies, improving its ability for empathetic user interactions. The chit-chat function aims to: 1) ease patients' nerves, 2) build trust for more intimate conversations, encouraging patients to disclose symptoms for accurate diagnosis, and 3) control dialogue flow and smoothly transition topics.

\begin{figure}[htbp]
   \centering
   \noindent\includegraphics[width=\columnwidth]{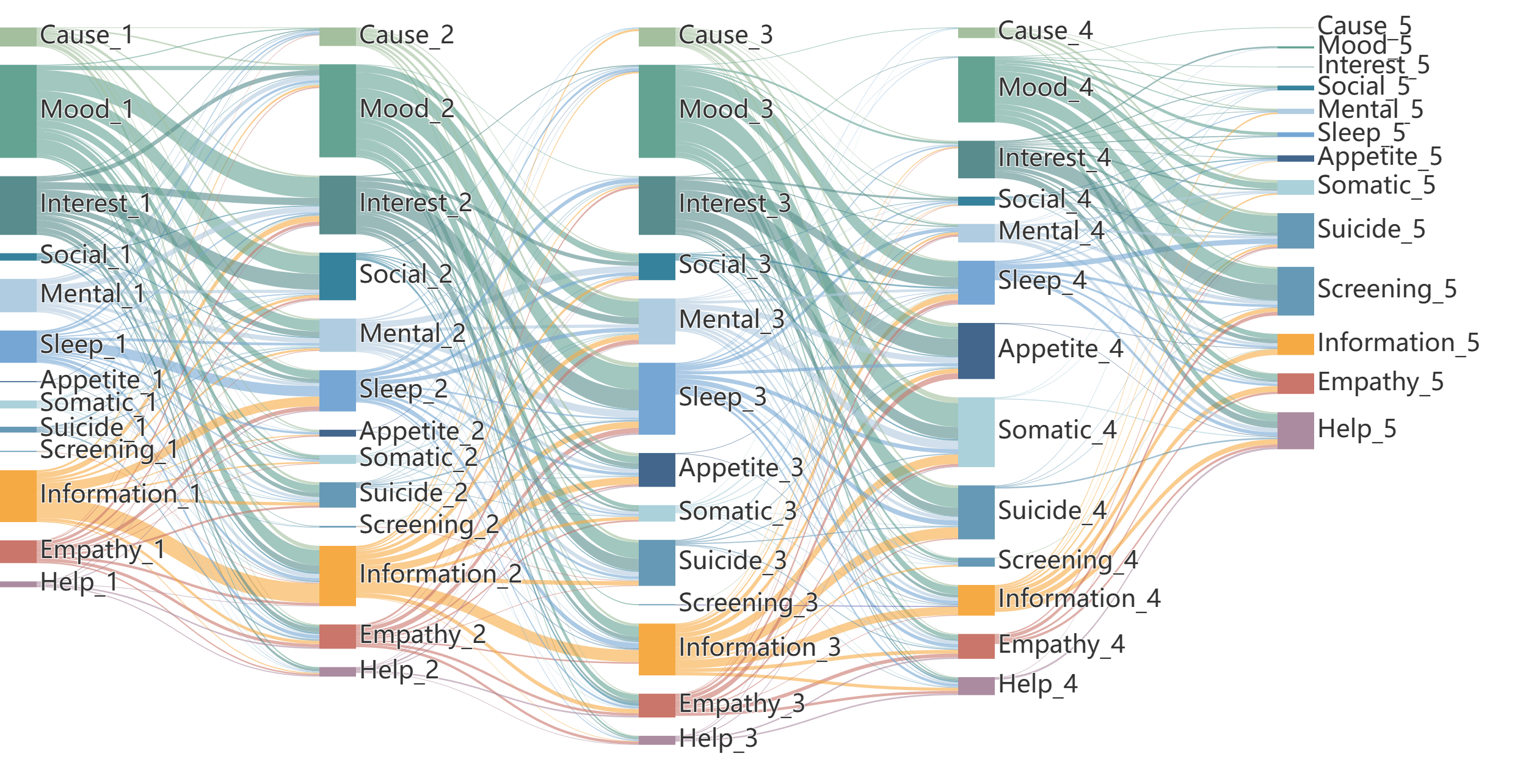}
   \caption{Intent Transition Tendency. The height represents the proportion of dialogues at this core symptoms and empathy strategies mentioned in \Scref{subsec:ontology definition} 
   }
   \label{fig:topic}
\end{figure}

\subsection{Ontology Annotation}

\paragraph{D$^4$ Introduction}
D$^4$~\cite{yao-etal-2022-d4} is a Chinese close-to-clinical-practice depression diagnosis dataset, containing 1,339 simulated dialogues conducted by well-trained patients and doctors. The statistics of D$^4$ are in \Cref{tab:d4 characteristics}.
\begin{table}[htbp]
    \centering
    \resizebox{1\linewidth}{!}{
    \begin{tabular}{cccc}
    \toprule
    \textbf{Categoty} & \textbf{Total} & \textbf{Patient} & \textbf{Doctor} \\
    \midrule
    \textbf{Dialoges} & 1,339 & - & - \\
    \textbf{Avg. turns} & 21.6 & - & - \\
    \textbf{Avg. utterances per dialogue} & 60.9 & 30.9 & 29.9 \\
    \textbf{Avg. tokens per dialogue} &  877.6 &  381.8 & 495.8
 \\
    \bottomrule
    \end{tabular}}
    \caption{Original statistics of D$^4$.}
    \label{tab:d4 characteristics}
\end{table}

\paragraph{Annotation Process}
We carefully designed our annotation process, which included: 1) \textbf{Manual Creation}: Developing a detailed manual for ontology definition, covering symptom-related and empathy-related ontologies. 2) \textbf{Annotator Training}: Training annotators with the manual, initially annotating 20 dialogues ourselves as a validation set, then assessing the annotators' alignment with the manual based on their annotation of this set. 3) \textbf{Annotation Process}: Professional annotators annotated dialogues, with periodic quality checks. They labeled the psychiatrist's intent and the patient's dialogue states. 4) \textbf{Automatic Labeling}: For ontology labeling, we used the BERT model as a classifier, with two separate classifiers for symptom-related and empathy-related ontologies due to their distinct nature. The intent labeling classifier's performance, indicating its effectiveness and accuracy, is reported in \Cref{tab:intent classification}. We employ a team of trained annotators who has related work experience to annotate one-third of D$^4$ dialogues based on SEO. Rigorous scrutiny is then applied to the labeled dialogues, followed by the training of a classifier using the labeled dialogues to label the remaining dialogues automatically. The examples of SEO annotated data is shown in \Cref{fig: annotation}.

\begin{table}[htbp]
    \centering
    \resizebox{1\linewidth}{!}{
    \begin{tabular}{cccc}
    \toprule
    \textbf{Ontology} & \textbf{F1-Score} & \textbf{Precision} & \textbf{Recall}\\
    \midrule
    TOD & 93.03 / 85.97 / 92.70 & 93.49 / 87.25 / 93.49 & 92.57 / 85.39 / 92.58\\
    Chit-chat & 78.23 / 79.44 / 77.63 & 76.62 / 78.09 / 76.44 & 79.91 / 81.61 / 79.91\\
    \bottomrule
    \end{tabular}}
    \caption{Performance of Ontology Classification.  The performance of the TOD surpasses that of Chit-chat, owing to its adept handling of the intricate linguistic diversity inherent in chit-chat interactions. The numerical values presented in each cell correspond to the computed micro/macro/weighted averages.
    }
    \label{tab:intent classification}
\end{table}

\subsection{Data Characteristics}
\label{subsec:data characteristics}

To gain insights into the statistical aspects of the D$^4$ dataset annotated by SEO, we comprehensively analyzed the intent distribution and intent selection tendency within the dataset. In the subsequent sections, we will delve into the intricacies of these characteristics, shedding light on the underlying trends and tendencies observed in the data.

\paragraph{Intent Distribution} 

The intent distribution of D$^4$ is visually depicted in \Cref{fig:sunburst} and quantitatively summarized in \Cref{tab:intent distribution}. The whole SEO contains 38 kinds of TOD ontology and seven kinds of Chit-chat ontology. Upon examining the tabular representation and sunburst visualization, it becomes apparent that the intent distribution between TOD and Chit-chat categories is relatively well-balanced. However, the definition of SEO is a fine-grained 3-level hierarchy multi-label task, which is imbalanced in the number in the case of specific third ontology intents. This non-uniform distribution challenges learning the patterns associated with sparse intents, thus increasing the complexity.

\begin{table}[htbp]
    \centering
    \resizebox{1\linewidth}{!}{
    \begin{tabular}{ccccc}
    \toprule
    \textbf{Intent Type} & \textbf{Slot Num.} & \textbf{Total Num.} & \textbf{Avg. intents}\\
    \midrule
    TOD & 38 & 21,412 &  0.73 \\
    Chit-chat & 7 & 14,157 & 0.48 \\
    \midrule
    Total & 45 &35,569 & 1.21 \\
    \bottomrule
    \end{tabular}}
    \caption{Ontology intent statistics of D$^4$. \textbf{Avg. intents} indicates the average intents number per utterance.}
    \label{tab:intent distribution}
\end{table}

\paragraph{Intent Selection and Transition Tendency} 
\begin{figure}[htbp]
   \centering
   \noindent\includegraphics[width=\columnwidth]{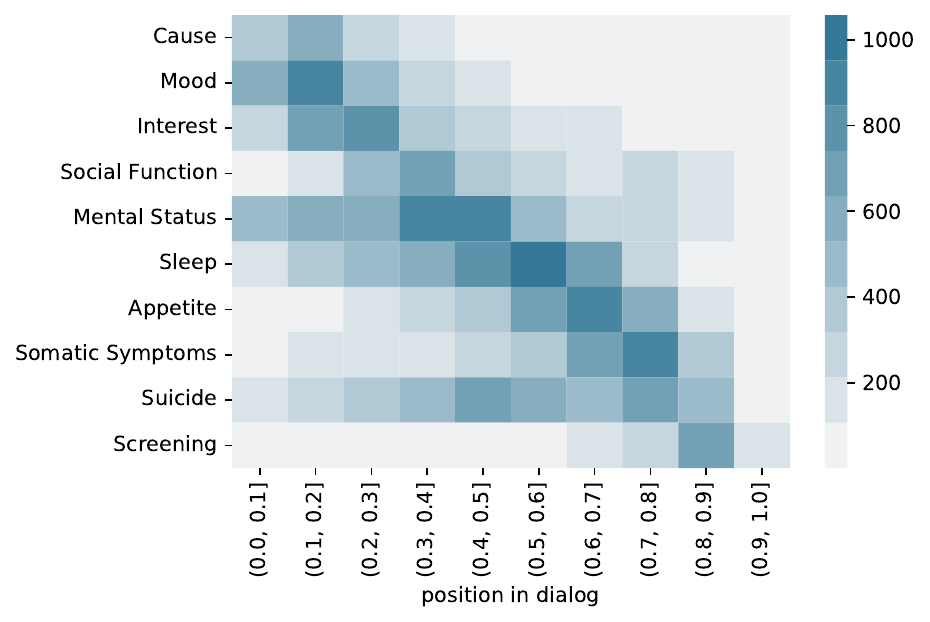}
   \caption{Symptom Intent Selection Tendency. The value on the vertical axis indicates the core depressive symptoms mentioned in \Scref{subsec:ontology definition}}
   \label{fig:heatmap}
\end{figure}

The tendency of intent selection in D$^4$ is graphically depicted in \Cref{fig:heatmap}, while the intent transition trend is illustrated in \Cref{fig:topic}, enabling us to discern a distinct pattern in the dialogue progression. Notably, the psychiatrist exhibits a discernible preference for inquiring about the user's mood and interest status as initial inquiries, followed by inquiries about social function, mental status, sleep, and appetite, eventually addressing suicide and somatic symptoms. In the meanwhile, the empathy-related chitchat strategies transition during the diagnosis process shows the construction of the therapeutic alliance: the doctor tends to get more personal information of the patient at first, and provide suggestion at the end of the diagnosis. Moreover, the intent transition trend provide us with a view of the relevance between symptoms and empathetic strategies. This discernible intent selection tendency indicates the potential for predicting intents in depression diagnosis dialogues. Such insights hold significant implications for enhancing the efficiency and effectiveness of the diagnostic process.

\subsection{Summary}

We can discern the challenges that arise during the processing of depression diagnosis dialogues. These challenges encompass the following aspects: 1) \textbf{Flexible Intent Transition}: \Cref{fig:heatmap} and \Cref{fig:topic} illustrates that while there is a discernible tendency in the intent selection, there is a lack of specific or relatively stringent guidelines to direct the choice of intents for subsequent utterances. Moreover, most utterances in the D$^4$ dataset comprise a combination of multiple Task-Oriented Dialog (TOD) intents and Chit-chat intents, which significantly complicates the task of intent prediction. 2) \textbf{Relatively Long Turns of Dialogues}: Notably, the average number of turns in D$^4$ dialogues is 21.55, surpassing that of conventional dialogue datasets such as MultiWOZ~\cite{budzianowski2018multiwoz}. Extended turns within dialogues contribute to a loss of dialogue history in the model's memory, leading to the potential repetition of previously asked questions and a decline in user engagement. 3) \textbf{Unbalanced yet Diverse Intents Distribution}: Our ontology definition comprehensively delineates depression diagnosis criteria. However, certain intents, despite their significance, occur infrequently, making it challenging for the model to discern their underlying patterns.

\section{Task Setup}
To address the challenges presented in \Scref{subsec:data characteristics}, we employ an intent prediction method to mitigate the challenges associated with flexible intent transitions. A dialogue state tracking method is also utilized to tackle the complexities arising from long-turn dialogues. Furthermore, we adopt a dialogue generation approach based on depression diagnosis intents to overcome the issue of unbalanced intent distribution, ensuring that all intents are adequately represented in the generated dialogues.

\paragraph*{Intent Prediction}

Our intent prediction task is to anticipate the intentions that a psychiatrist is likely to adopt in their forthcoming utterances based on the dialogue history. Since SEO framework provides a comprehensive three-level definition, as illustrated in \Cref{fig:sunburst}, wherein intents belonging to the same categories demonstrate significant similarities. While the classification model may yield accurate results, the generation model holds the potential to generate dialogue flows that capture the nuances of semantic meaning. Therefore, combining these models can contribute to a more comprehensive and effective intent prediction system. In order to enhance the performance of intent prediction, we explore three distinct approaches: 1) \textbf{Classification Model}: In this approach, we treat intent prediction as a multi-label classification task. By leveraging the classification model, we aim to assign multiple intent labels to each utterance accurately. 2) \textbf{Generation Model}: Another approach treats the task as a sequence generation task. By employing a generation model, we enable the generation of intent labels directly 3) \textbf{Extracting Predicted Intents from Response Generation}: Given that we employ an end-to-end generation model for response generation, we can extract the intents predicted by the response generation model itself. This approach lets us obtain the predicted intents directly from the generated responses, thus facilitating a seamless integration between response generation and intent prediction. To further improve the performance of intent prediction, we adopt an intent infusion mechanism, which considers the task's precision and recall, respectively. By combining the predictions from multiple models, we aim to leverage the classification model's precision and the generation model's semantic meaning, resulting in more robust and accurate intent predictions. This infusion approach enhances the overall performance and reliability of intent prediction.

\paragraph*{Dialogue State Tracking}
\label{subsec:dialogue state tracking}
Dialogue state tracking is crucial in summarizing the current symptoms inferred from dialogue history. This process is vital in monitoring users' mental health states and determining the subsequent question intents. However, regarding dialogue state tracking, we encounter a challenge wherein the number of dialogue states to be tracked exceeds the scope of conventional dialogue state tracking tasks~\cite{budzianowski2018multiwoz}. Consequently, if we were to track the current dialogue state through the entire history, the task would become more difficult due to the sparsity of dialogue state changes. To overcome this challenge, we employ an alternative dialogue state tracking strategy, referred to as \textbf{Res-Track}. The function is explained in \Cref{eq:res-track}, where $DS_k$ indicates the user's dialogue state of the $k$-th psychiatrist's utterance, $DH_{\{k-3:k\}}$ is the last three turns of dialogue history. This strategy supplements the conventional classifier by solely detecting changes in dialogue states, thereby streamlining the tracking process.

\begin{equation}
    \begin{aligned}
        &\textbf{Res-Track}(DS_{k-1};DH_{\{k-3:k\}}) = \Delta DS_k\\
        &= DS_k - DS_{k-1}
    \end{aligned}
    \label{eq:res-track}
\end{equation}

\paragraph*{Response Generation}
\label{subsec:depression diagnosis intents}

Apart from training an end-to-end response generation model like previous work D$^4$, we utilize predicted intents to overcome the relatively low intent accuracy. In order to explore the influence of predicted intents on response generation, we employ two strategies for this task: 1) \textbf{Encoder Method}: In this approach, the dialogue history and the intents that the psychiatrist is expected to adopt are provided to the generation model. The model then generates a response from the perspective of the psychiatrist. This method emphasizes the relationship between the intents and the corresponding response. 2) \textbf{Decoder Method}: Here, the predicted intents of the psychiatrist serve as the generated intents. This strategy focuses on both intent prediction and response generation. By adopting decoder method, we aim to explore the interplay between intent prediction and subsequent response generation.

\section{Experiments}

We experimented with different strategies with state-of-the-art pre-trained language models CPT (121M)~\cite{shao2021cpt} and BERT (110M)~\cite{devlin2019bert} as our backbone models, the same as prior works~\cite{yao-etal-2022-d4}. The specific hyperparameter settings are provided in Appendix \Cref{tab:hyperparameters}. Moreover, considering the promising capabilities of Large Language Models, we also utilized LLMs such as GPT-3.5 (gpt-3.5-turbo-0613, to be specific) to conduct our experiments. 

The experiments include 4 classic dialogue tasks: \textbf{\textit{intent prediction}} aims to predict the intents of the next utterance psychiatrist would like to utilized based on the conversation history, \textbf{\textit{dialogue state tracking}} concludes symptoms the patients had we can infer from the context, \textbf{\textit{response generation}} in order to generate psychiatrists’ probable response based on the dialog, \textbf{\textit{summary generation}} generates symptom summaries based on the entire dialog history and dialogue states, and depression-related task \textbf{\textit{risk classification}}, which predicts the severity of depressive episodes based on the dialogue context and dialogue summary. In the meanwhile, the performance of LLMs based on unsupervised methods is much worse than that of the supervised methods of trainable size models because of the specialization and diversity of ontology definition in this scenario. Therefore, even though the SEO is also found beneficial to LLMs, we attach the experiments result of LLMs in Appendix \Cref{tab: gpt response generation} serving as a supplementary reference. 
 \begin{table}[htbp]
    \centering
    \resizebox{1\linewidth}{!}{
    \begin{tabular}{cccc}
    \toprule
        \textbf{Classifier} & \textbf{F1-Score} & \textbf{Precision} & \textbf{Recall}\\
    \midrule
    CPT~(Generate) & 35.11 & 36.79 & 33.59 \\
    BERT~(Classify) & 38.73 & 48.00 & 32.46 \\
    CPT~(Extract) & 37.70 & 38.93 & 36.54 \\
    \midrule
    Infusion (Recall) & 38.13 & 30.61 & \textbf{50.53} \\
    Infusion (Precision) & \textbf{40.44} & \textbf{53.71} & 32.43 \\
    \bottomrule
    \end{tabular}}
    \caption{Performance of Intent Prediction. The recall infusion model merges the intents extracted from three foundational models, while precision infusion model adopts a voting approach to optimize precision.}
    \label{tab:intent prediction}
\end{table}

\begin{table}[htbp]
    \centering
    \resizebox{1\linewidth}{!}{
    \begin{tabular}{cccc}
    \toprule
    \textbf{Classifier} & \textbf{JGA} & \textbf{SA} & \textbf{Symptom F1}\\
    \midrule
    BERT~(Baseline) & 19.60 & 92.74 & 72.63 \\
    BERT~(Res-Track) & \textbf{29.05} & \textbf{94.59} & \textbf{76.81} \\
    \bottomrule
    \end{tabular}}
    \caption{Dialogue State Tracking Performance. Baseline model tracks the entire dialogue state throughout the dialogue process, while Res-Track model solely detects the change in dialogue state.}
    \label{tab:dialogue state tracking}
\end{table}
\subsection{Evaluation Metrics}
Our evaluation metrics are three-fold: 1) \textbf{Classification metrics}, which are utilized for classifying tasks such as intent classification, dialogue state tracking, emotion recognition, and severity classification. The classification metrics include accuracy, precision, recall, F1-Score, and so on. 2) \textbf{Generation metrics}, used for generation tasks such as response generation. The generation metrics consists of BLEU-2~\cite{papineni2002bleu}, Rouge-L~\cite{lin2004rouge}, METEOR~\cite{banerjee2005meteor} and DIST-2~\cite{li2015diversity} to measure the generation quality. 3) \textbf{Engagement metrics}, designed to measure the model's performance in engagement. The metrics are inspired by previous work~\cite{chen2023llmempowered,deng-etal-2023-knowledge}. These metrics include \textit{in-depth questions ratio}, \textit{repeated questions ratio}, and \textit{empathy response ratio}. The design of engagement metrics is presented in Appendix \Scref{sec:engagement metrics}.

\begin{table*}[htbp]
    \centering
    \resizebox{1\linewidth}{!}
    {
    \begin{tabular}{ccccccccc}
        \toprule
        \textbf{Information Provided} & \textbf{Ontology} & \textbf{Intents} &\textbf{Intents Provide Method}  & \textbf{BLEU-2} & \textbf{ROUGE-L} & \textbf{METEOR} & \textbf{DIST-2} \\
        \midrule
        \multirow{3}{*}{--} & Baseline & No intent & -- & 19.79 & 0.36 & 0.2969 & 0.07 \\
        \cmidrule{2-8}
        & \multirow{1}{*}{D$^4$} & \multirow{1}{*}{5 Intents} & -- & 20.43 & \textbf{0.38} & 0.3005 & 0.08 \\
        \cmidrule{2-8}
        & \multirow{3}{*}{SEO}  & \multirow{3}{*}{45 Intents} & -- & 20.92 & 0.36 & 0.3039 & 0.13 \\
        \multirow{2}{*}{Predicted Intents}& & & Decoder & \textbf{21.28} & 0.36 & 0.3159 & 0.13\\
        & & & Encoder & 20.37 & 0.35 & \textbf{0.3183} & \textbf{0.14}\\
        \midrule
        \multirow{4}{*}{Golden Intents} & \multirow{2}{*}{D$^4$} & \multirow{2}{*}{5 Intents} & Decoder & 22.61 & 0.39 & 0.3261 & 0.08 \\
        & & & Encoder & 23.06 & 0.39 & 0.3328 & 0.10 \\
        \cmidrule{2-8}
        & \multirow{2}{*}{SEO} & \multirow{2}{*}{45 Intents} & Decoder & 28.84 & 0.45 & 0.3991 & 0.14 \\
        & & & Encoder &\textbf{30.54} & \textbf{0.47} & \textbf{0.4227} & \textbf{0.15} \\
        \bottomrule
    \end{tabular}}
    \caption{Model Performance of Response Generation, where D$^4$ model only consider 5 topics as intents. Encoder/Decoder = whether and how we provide intents to the model. Decoder only uses the label in the inference stage, while the encoder is in the training stage. Predicted intents are obtained via the recall infusion method.}
    \label{tab:response generation}
\end{table*}
\subsection{Experimental Results}

\paragraph{Intent Prediction}
\label{subsec:annotation task}
 In this task, we employ BERT as the classification backbone model for our intent prediction. The performance of intent prediction is comprehensively illustrated in \Cref{tab:intent prediction}. The findings demonstrate the efficacy of the infusion model, thereby offering promising prospects for improved response generation performance.

\paragraph{Dialogue State Tracking}
 We conducted a comparative analysis between two dialogue state tracking classifiers: the baseline classifier, which detects the entire dialogue state, and the \textbf{res-track} classifier, which focuses explicitly on capturing dialogue state changes as discussed in \Scref{subsec:dialogue state tracking}. Both classifiers were trained with BERT. The experimental results are summarized in \Cref{tab:dialogue state tracking}. To comprehensively evaluate the performance of the dialogue state tracking models, we employed two evaluation metrics: \textbf{Joint Goal Accuracy} (JGA)~\cite{8268986} and \textbf{Slot Accuracy} (SA). Apart from the traditional DST evaluation metrics, we proposed a novel evaluation metric called \textbf{Symptom F1}, which evaluates the F1-Score of the TOD-related dialogue state in the final turn, suggesting the symptom accuracy of the DST model. The results indicate that the Res-Track strategy outperforms the baseline classifier, demonstrating its superior efficiency and effectiveness in tracking dialogue state changes.

\paragraph*{Response Generation}

Our evaluation of response generation focuses on both generation and engagement metrics using CPT, with results in \Cref{tab:response generation} and \Cref{tab:engagingness metrics}. Key observations include: 1) The SEO-annotated D$^4$ model, with its 45 distinct intents, outperforms the original D$^4$ that considered only five topics as intents, demonstrating SEO's enhanced response generation capabilities. 2) The decoder method(utilizing intent during inference) is more effective than the encoder method(utilizing intent during training) when no other information(golden intents) provided. This indicates the decoder's focus more on response generation considering the context. In the meanwhile, the encoder method excels decoder method with golden intent, suggesting a superior response generation approach with precise intents. 3) The SEO approach shows improved engagingness over the original D$^4$, highlighting the benefits of incorporating empathy strategies into response generation for more compelling conversations. A detailed case analysis in Appendix \Scref{sec:case} further supports these findings.

\begin{table}[htbp]
    \centering
    \resizebox{1\linewidth}{!}{
    \begin{tabular}{ccccc}
        \toprule
        \textbf{Ontology Type} & \textbf{Repeat Ratio} & \textbf{Empathy Ratio} & \textbf{In-depth Ratio} \\
        \midrule
        Reference & 2.36 & 32.38 & 24.60\\
        \midrule
        Baseline & 3.94 & 2.71 & 19.7\\
        D$^4$ & 6.69 & 6.02 & 18.73 \\
        SEO & \textbf{2.64} & \textbf{12.25} & \textbf{22.18}\\
        \bottomrule
    \end{tabular}}
    \caption{Model Performance of Response Generation on Engagement Metrics. The reference refers to engagement scores we evaluated on the testset. The SEO refers to the Decoder method based on its better performance in automatic metrics. The description of the engagement metrics is presented in Appendix \Scref{sec:engagement metrics}. }
    \label{tab:engagingness metrics}
\end{table}

\paragraph*{Summary Generation}

We use two different methods for summary generation task: 1) We implement the same summary generation method in D$^4$, taking the dialog history solely as input. 2) We append final dialogue state to the end of the dialogue history, providing a holistic perspective of the user. The experimental outcomes are presented in Table \ref{tab:summary generation}. Notably, our finding indicates that including dialogue state contributes to summary generation.

\begin{table}[ht]
    \centering
    \resizebox{1\linewidth}{!}{
    \begin{tabular}{ccccc}
        \toprule
        \textbf{Input Type} & \textbf{BLEU-2} & \textbf{ROUGE-L} & \textbf{METEOR} & \textbf{DIST-2} \\
        \midrule
        Dialog & 0.14 & 0.22 & 0.2122 & 0.16\\
        Dialog + State & \textbf{0.15} & \textbf{0.28} & \textbf{0.2218} & \textbf{0.20}\\
        \bottomrule
    \end{tabular}}
    \caption{Model Performance of Summary Generation.}
    \label{tab:summary generation}
\end{table} 

\paragraph*{Risk Classification}

\begin{table}[ht]
    \centering
    \resizebox{1\linewidth}{!}{
    \begin{tabular}{ccccccc}
        \toprule
        \multirow{2}{*}{\textbf{Depression Severity}} & \multicolumn{3}{c}{\textbf{D$^4$ (Dialog)}} & \multicolumn{3}{c}{\textbf{SEO (Dialogue State)}} \\
        & \textbf{Precision} & \textbf{Recall} & \textbf{F1-Score} & \textbf{Precision} & \textbf{Recall} & \textbf{F1-Score}\\
        \midrule
        None & 0.74 & 0.60 & 0.66 & 0.74 & 0.72 & 0.73\\
        Mild & 0.43 & 0.46 & 0.41 & 0.47 & 0.21 & 0.29\\
        Moderate & 0.36 & 0.38 & 0.35 & 0.41 & 0.67 & 0.51\\
        Severe & 0.31 & 0.36 & 0.32 & 0.53 & 0.47 & 0.50\\
        \midrule
        \textbf{Average} & 0.49 & 0.47 & 0.46 & \textbf{0.55} & \textbf{0.54} & \textbf{0.52}\\
        \bottomrule
    \end{tabular}}
    \caption{Model Performance of 4-class Depression Risk Classification, Spanning from \textit{None, Mild, Moderate and Severe} Depressive Risk.}
    \label{tab:risk classification}
\end{table}

We employ 4-class depression risk classification approach using CPT in evaluation. While prior studies utilized dialogue history as input, we differentiate ourselves by using informative dialogue states. The result of our experiment are summarized in \Cref{tab:risk classification}, and the result shows dialogue states leads to superior performance. This observation indicates the significance and utility of dialogue states in the context of depression risk classification.

\subsection{Human Evaluation}

To comprehensively assess our model's generative performance, we conducted human evaluation focusing on \textbf{fluency} (the response's clarity), \textbf{coherence} (consistency with dialogue history and topic transition), and \textbf{empathy} (understanding the patient's feelings). Evaluators rated 30 responses, including ground truth, responses from our D$^4$ model, and SEO model responses, on a 1-3 scale, the detailed scale definition has shown in \cref{tab:human evaluation scale}. The evaluation involved 20 volunteers aged 22-56, all with at least a bachelor's degree, comprising 60\% male and 40\% female participants. Result, detailed in \Cref{tab:human evaluation}, shows the SEO model excelling in all aspects, aligning with our experimental findings.

\begin{table}[ht]
    \centering
    \resizebox{1\linewidth}{!}{
    \begin{tabular}{cccc}
        \toprule
        \textbf{Ontology Type} & \textbf{Fluency} & \textbf{Coherence} & \textbf{Empathy} \\
        \midrule
        Reference & 2.74 & 2.36 & 2.32 \\
        \midrule
        D$^4$ & 2.68 & 2.04 & 2.09 \\
        SEO & \textbf{2.71} & \textbf{2.22} & \textbf{2.16} \\
        \bottomrule
    \end{tabular}}
    \caption{Human Evaluation Performance on Reference, D$^4$, and SEO. The reference refers to the ground truth.}
    \label{tab:human evaluation}
\end{table}

\section{Related Works}
This section focus on intersection of task-oriented and chit-chat dialogues in depression diagnosis, discussing advancements in mental health chatbots.
\paragraph*{Task-Oriented Chat}
There have been attempts to combine task-oriented and chit-chat dialogue, but these approaches relied on combining task-oriented and chit-chat dialogues~\cite{sun2020adding, young2022fusing} rather than conversations in real-world scenarios. There have also been attempts to incorporate chit-chat as a specialized action within existing task-oriented dialogue ontologies~\cite{zhao2021unids, yao-etal-2022-d4}. Nevertheless, they are too coarse to handle the nuances of depression diagnosis dialogues, which involves a larger proportion and more complex emotional support.
\paragraph*{Mental Health Chatbots}
Chatbots are instrumental in large-scale mental disorder screening, especially for early depression detection. They mitigate discomfort in traditional diagnostics, posing questions from scales with users responding to specific options~\cite{Philip_2017,10.1145/3308532.3329469}. Recent models allow more personalized responses with integrated semantic understanding modules~\cite{arrabales2021perla}. Previous studies outlined complex diagnostic systems for mental disorders~\cite{10.5555/2615731.2617415}, but combining accurate diagnostics with empathy in automated systems is challenging. Researchers are exploring deep reinforcement learning to enhance empathetic responses in telemedicine~\cite{sharma2021facilitating}. 

\section{Conclusion}
Task-oriented chat is an interesting yet complex dialogue type aiming at optimizing TOD and chit-chat while being less explored for lack of real-world scenarios. This work situates itself as a first tryout to investigate TOC in depression-diagnosis-chat by proposing a joint ontology definition and generation framework, combining task-oriented conversations' reliability with empathy's appeal in depression diagnosis. SEO ontology enables intent transition tracking, vital for accurate and empathetic depression diagnosis, and serves as an inspiring framework for TOC in other domains. Exhaustive experimental results indicate significant improvements in task completion and emotional support in depression diagnosis, fostering a more comprehensive approach to TOC dialogue system development and applications in digital mental health.

\section{Limitations}
Our research has limitations that can be further improved in future work.
\begin{enumerate}
    \item \textbf{Empathy Quality: }Although our model can reasonably predict empathy-related labels, it tends to exhibit issues such as grammar confusion and incomplete sentences when generating empathetic statements. For example, a common error is the incorrect use of pronouns. Upon analysis, we believe a significant contributing factor is the model's lack of common-sense information, resulting in a shallow understanding of certain semantics. Consequently, it struggles to extract and utilize information effectively from the preceding context. By introducing a trial system, we may have an opportunity to improve this situation.
    \item \textbf{Evaluation Metrics: }
    During the training of our model, the BLEU score served as the optimization metric. However, we observed a notable disparity upon conducting a distinct evaluation of response generation. Specifically, we noticed that the BLEU score for empathy-related sentences was significantly lower than sentences containing only TOD labels
    . This discrepancy can be attributed to empathetic statements' inherent complexity and diversity, which often lack a definitive, objectively correct answer. As a result, conventional n-gram metrics possess significant limitations when accurately assessing the precision and linguistic richness of empathetic sentences.
    \item \textbf{Utilization of Labels: }In the model training, we used patient states and doctor intents as data inputs. However, we also labeled patient intents and emotions during the annotation process, which were not utilized in subsequent tasks. In future work, we may be able to leverage patient-related information to enhance the quality of empathy further, obtain more targeted empathetic strategies, and achieve more precise risk predictions.
\end{enumerate}

\section{Ethical Considerations}
When using non-public datasets, we store and process the data in a reasonable and compliant manner, taking appropriate security measures to prevent unauthorized access and data breaches while ensuring that the data is only used for research related to the diagnostic system to prevent data misuse. To protect the providers of the portraits, we remove relevant details that may lead to the exposure of their real identities, ensuring the anonymity of the data.

During the data annotation process, we provide fair and reasonable compensation to all annotators and actively communicate with them to prevent any potential psychological harm from prolonged exposure to conversations containing negative emotions.

In the model's training process, the sole criterion for selecting dialogue samples is the objective quality of the data, unaffected by factors such as gender, race, age, or region, which ensures the model can provide accurate, neutral, and fair results.

In the human evaluation process, we provide fair compensation to all volunteers and ask them to sign a confidentiality agreement to ensure that the dataset information is not leaked.

\bibliography{custom}

\appendix



\section{Model Training Setting}

The experimental framework consists of two trainable models: BERT and CPT. The implementation of the experiment is fine-tuned through a set of meticulously chosen hyperparameters, outlined comprehensively in \Cref{tab:hyperparameters}. Our methodology adheres closely to the configuration employed in the antecedent study, D$^4$~\cite{yao-etal-2022-d4}, thereby establishing a sense of continuity and enabling direct comparative analysis. Our model was trained utilizing single Nvidia 2080Ti, with 9 hours in average to train the model.

\begin{table}[htbp]
    \centering
    \resizebox{1\linewidth}{!}{
    \begin{tabular}{ccc}
    \toprule
    Hypermarameter & BERT & CPT \\
    \midrule
    learning rate & 1e-5 & 1e-5 \\
    weight decay & 1e-6 & 1e-6 \\
    learning rate scheduler type & cosine & cosine \\
    warmup steps & 100 & 100 \\
    seed & 42 & 42 \\
    max length (when generate sequence) & -- & 512\\
    \bottomrule
    \end{tabular}}
    \caption{Hyperparameters Utilized in The Training Stage.}
    \label{tab:hyperparameters}
\end{table}

\section{Engagement Metrics}
\label{sec:engagement metrics}

The engagement metrics contains repeated ratio, in-depth questions ratio, empathy ratio. The \textit{In-depth Questions Ratio} is directly inspired by previous LLMs evaluation work~\cite{chen2023llmempowered}, which indicates the behavior of asking follow-up questions about the patient’s symptom. Since the evaluation work just consider the core symptoms, it is not enough to evaluate the chatbots we utilized if they ask the same question all the time, which will harm the engagingness of the patient. Therefore, we come up with \textit{Repeated Questions Ratio} which is inspired by previous emotional support work~\cite{deng-etal-2023-knowledge}, which implies the diversity of the chatbots, measuring whether they will ask the exact same questions. At last, we purpose \textit{Empathy Ratio}, which evaluate the proportion of the empathetic response of the outputs. While other empathy metrics~\cite{sharma-etal-2020-computational} mainly consider empathy of the response generated by the model, we evaluate whether the model can generate appropriate empathetic response considering the dialogue history.

\paragraph{In-depth Questions Ratio} In-depth questions ratio (IQR) is calculated as \Cref{eq:IQR}, where $\mathcal{IQ}_{\{n,k\}}$ indicates the in-depth questions set of the $k$-th psychiatrist's utterance of the $n$-th dialogue; $T_{\{n,k-i:k\}}$ means the topics (core symptoms, which are defined in \Scref{subsec:ontology definition}) mentioned between $(k-i)$-th and $k$-th psychiatrist's utterance of the $n$-th dialogue, $i$ is the hyperparameter of the screening window, which we set 3 and 5 in the experiments.
\begin{equation}
    \begin{aligned}
       IQR = & \frac{\sum^{\mathcal{N}}_{n=1}\sum^{\mathcal{L}_n}_{k=1}\lvert\{x|x\in \mathcal{IQ}_{\{n,k\}}\}\rvert}{\sum^{\mathcal{N}}_{n=1} \mathcal{L}_n}\\
       \mathcal{IQ}_{\{n,k\}} = & \{x|x\in I_{\{n,k\}},x\notin I_{\{n,1: k-1\}},\\
             & x\in T_{\{n,k-i:k\}}\}
    \end{aligned}
    \label{eq:IQR}
\end{equation}

\paragraph{Repeated Questions Ratio} Repeated questions ratio (RQR) is calculated as \Cref{eq:RQR}, where $\mathcal N$ indicates the number of dialogues in the dataset; $\mathcal{L}_n$ is the utterance length by psychiatrist of the dialogue $n$; $\mathcal{R}_{\{n,k\}}$ means the repeated questions set; $I_{\{n,k\}}$ means the intent set of the $k$-th psychiatrist's utterance of the $n$-th dialogue; $I_{\{n,k-i:k\}}$ is the union intent set between $(k-i)$-th and $k$-th psychiatrist's utterance of the $n$-th dialogue. $DS_{\{n,k\}}$ indicates the dialogue state of the $k$-th psychiatrist's utterance of the $n$-th dialogue, $i$ is the hyperparameter of the screening window, which we set 3 the experiments.
\begin{equation}
    \begin{aligned}
       RQR = & \frac{\sum^{\mathcal{N}}_{n=1}\sum^{\mathcal{L}_n}_{k=1}\lvert\{x|x\in \mathcal{RQ}_{\{n,k\}}\}\rvert}{\sum^{\mathcal{N}}_{n=1} \mathcal{L}_n}\\
       \mathcal{RQ}_{\{n,k\}} = &  \{x|x\in I_{\{n,k\}},x\in I_{\{n,k-i: k-1\}},\\
          &  x\in DS_{\{n,k-1\}}\}
    \end{aligned}
    \label{eq:RQR}
\end{equation}

\paragraph{Empathy Response Ratio} Empathy Ratio (ERR) is calculated as \Cref{eq:ER}, where $\mathcal{E}_{\{n,k\}}$ indicates the empathy set of the $k$-th psychiatrist's utterance of the $n$-th dialogue; $ES$ means the empathy strategies set we defined in the ontology manual in Sec. \ref{subsec:ontology definition}.
\begin{equation}
    \begin{aligned}
       ERRR = & \frac{\sum^{\mathcal{N}}_{n=1}\sum^{\mathcal{L}_n}_{k=1}\mathbb{I}(\mathcal{E}_{\{n,k\}} \neq \varnothing)}{\sum^{\mathcal{N}}_{n=1} \mathcal{L}_n}\\
       \mathcal{E}_{\{n,k\}} = & \{x|x\in I_{\{n,k\}},x\in ES\}
    \end{aligned}
    \label{eq:ER}
\end{equation}

\section{LLMs Experiment Result}

We utilized GPT-3.5 (to be specific, the gpt-3.5-turbo-0613) as the representative LLMs to conduct the experiments. We conducted our experiment of LLMs on \textbf{\textit{response generation}} (\Cref{tab: gpt response generation}),  and depression-related task \textbf{\textit{risk classification}} (\Cref{tab:gpt risk classification}), the prompt we utilized are shown in \Cref{tab:prompt}. The experiment result shows the effectiveness of SEO. We also conducted our experiment on ontology classification, intent prediction, and dialogue state tracking tasks. However, given the specialization and diversity of ontology definition in this scenario, LLMs often struggle to comprehend the intent requirements within the prompts, resulting in generating labels that do not meet the criteria. Therefore, the performances of LLMs on these tasks are not represented in these tasks.

\begin{table}[htbp]
    \centering
    \resizebox{1\linewidth}{!}{
    \begin{tabular}{ccccccc}
        \toprule
        \textbf{Ontology} & \textbf{Intents}  & \textbf{BLEU-2} & \textbf{ROUGE-L} & \textbf{METEOR} & \textbf{DIST-2} \\
        \midrule
        Baseline & No intent & 2.56 & \textbf{0.11} & 0.1180 & 0.10 \\
        D$^4$ & 5 Intents & 2.70 & \textbf{0.11} & 0.1245 & \textbf{0.12} \\
        SEO  & 45 Intents & \textbf{2.73} & \textbf{0.11} & \textbf{0.1321} & \textbf{0.12}\\
        \bottomrule
    \end{tabular}}
    \caption{Model Performance of Response Generation Based on GPT-3.5, where D$^4$ model only consider 5 topics as intents. }
    \label{tab: gpt response generation}
\end{table}

\begin{table}[ht]
    \centering
    \resizebox{1\linewidth}{!}{
    \begin{tabular}{ccccccc}
        \toprule
        \multirow{2}{*}{\textbf{Depression Severity}} & \multicolumn{3}{c}{\textbf{D$^4$ (Dialog)}} & \multicolumn{3}{c}{\textbf{SEO (Dialogue State)}} \\
        & \textbf{Precision} & \textbf{Recall} & \textbf{F1-Score} & \textbf{Precision} & \textbf{Recall} & \textbf{F1-Score}\\
        \midrule
        None & 1.0 & 0.02 & 0.05 & 0.93 & 0.21 & 0.34\\
        Mild & 0.07 & 0.06 & 0.06 & 0.32 & 0.18 & 0.23\\
        Moderate & 0.26 & 0.56 & 0.35 & 0.18 & 0.11 & 0.14\\
        Severe & 0.40 & 0.53 & 0.45 & 0.23 & 1.0 & 0.38\\
        \midrule
        \textbf{Average} & \textbf{0.47} & 0.25 & 0.19 & 0.46 & \textbf{0.29} & \textbf{0.26}\\
        \bottomrule
    \end{tabular}}
    \caption{Model Performance of 4-class Depression Risk Classification Based on GPT-3.5, Spanning from \textit{None, Mild, Moderate and Severe} Depressive Risk.}
    \label{tab:gpt risk classification}
\end{table}

\begin{table*}[htbp]
    \centering
    \begin{tabular}{m{0.2\linewidth} m{0.8\linewidth}}
    \toprule
    Task & Prompt \\
    \midrule
    Response Generation & I will have you play the role of a compassionate psychologist, conducting an inquiry with a visitor who might be suffering from depression. I will provide a segment of dialogue history with intention labels, where the doctor's statements begin with \textless doc\_bos\textgreater, the patient's statements begin with \textless pat\_bos \textgreater, and the doctor's intentions begin with \textless doc\_act \textgreater. Based on the conversation history and intention labels, I will ask you to anticipate what the doctor should say next and generate a response that starts with \textless doc\_bos \textgreater. Here's the conversation history:\\
    \midrule
    Depression Risk Classification (From Dialogue) & I will provide you with a description of a patient's dialogue regarding their symptoms. Please determine if the patient has depression. If the patient does not show signs of depression, choose [None]. If there is a tendency towards depression, based on the number and severity of symptoms, please determine the level of depression: [Mild], [Moderate], [Severe]. You are only allowed to choose one of these four options.\\
    \midrule
    Depression Risk Classification (From State) & I will provide you with a list of symptoms for the patient. Each \textless pat\_act \textgreater represents a symptom that the patient experiences. If a symptom is not mentioned, it means the patient doesn't have that issue. Please determine if the patient has depression. If the patient does not exhibit signs of depression, choose [None]. If there is a tendency towards depression, please assess the level of depression based on the number and severity of symptoms: [Mild], [Moderate], [Severe]. You are only allowed to select one of these four options. \\
    \bottomrule
    \end{tabular}
    \caption{Prompts Utilized in The Experiments.}
    \label{tab:prompt}
\end{table*}

\section{Definition of Human Evaluation Scale}
The detailed definition of Human Evaluation Scale is shown in \cref{tab:human evaluation scale}.
\begin{table*}[htbp]
    \centering
    \resizebox{1\linewidth}{!}
    {\begin{tabular}{ccc}
    \toprule
    Aspect & Scale & Definition \\
    \midrule
    \multirow{3}{*}{Fluency} & 1 & The answer is difficult to understand, and the sentence structure itself is not fluent enough.\\ 
    \cmidrule{2-3}
    & 2 & The answer is understandable, but the sentence structure feels uncommon or awkward.\\ 
    \cmidrule{2-3}
    & 3 & The answer is fluent and natural, with no comprehension issues.\\
    \midrule
    \multirow{3}{*}{Coherence} & 1 & The answer completely disregards the preceding text, exhibiting repetitive questioning behavior.\\
    \cmidrule{2-3}
    & 2 & The answer is average, with no issues in transitioning topics.\\ 
    \cmidrule{2-3}
    & 3 & The answer considers the conversation's history, with natural topic transitions.\\
    \midrule
    \multirow{3}{*}{Empathy} & 1 & The conversation is stiff and cold, with actions that may even offend the patient.\\
    \cmidrule{2-3}
    & 2 & There's no empathy in the conversation, following a normal conversational flow.\\ 
    \cmidrule{2-3}
    & 3 & The conversation demonstrates empathy, with attempts to understand the patient's emotional tendencies.
\\
    \bottomrule
    \end{tabular}}
    \caption{The Definition of Human Evaluation Scales}
    \label{tab:human evaluation scale}
\end{table*}

\section{Case Examples}
\label{sec:case}

We further provide a case analysis for a more intuitive comparison (\Cref{tab:case_anapysis}). The example has been translated from Chinese into English for better understanding.
SEO-based models exhibit several notable advantages compared to the previous D4 model: 1) \textbf{Accurate Intent Prediction}. In this example, the D4 model successfully completed the task of inquiring about the family medical history, even without predicting the symptom-related labels. However, despite predicting empathetic-related labels, it failed to exhibit any empathetic response. Our model, on the other hand, accurately performed label prediction and successfully completed the relevant task in subsequent dialogue generation without any deviation. 2) \textbf{Enhanced Empathetic Expression}. The results demonstrate that our models employ a targeted empathetic strategy by effectively extracting and utilizing relevant information of the patient's situation from the preceding context, and they can generate answers that are very close to real conversational data. At the same time, we have observed that the basic SEO model slightly falls behind SEO with golden encoder in terms of the richness and fluency of empathetic statements, but it still outperforms the previous D$^4$ model.

\begin{table*}[htbp]
    \centering
    \begin{tabular}{|p{15.5cm}|}
    \hline
    \textbf{History} 
    \par \textit{doctor}: May I ask if something happened recently?
    \par \textit{patient}: My ex-husband and I had a bad relationship, and we peacefully divorced over a month ago. However, we agreed to co-parent our child and keep the fact that mommy and daddy have separated a secret from him, in order to give him a complete childhood. But it seems like the child has become very sensitive. He keeps asking me if I divorced his father and even says that if we are truly divorced, he will run away from home. I'm really worried and don't know how to handle the situation with my child. It's all my fault for not being able to provide a complete family for the child.
    \par ...
    \par \textit{doctor}: Hmm, in this situation, have you talked to your family or friends about it? Have they given you any good advice or suggestions?
    \par \textit{patient}: I have discussed this issue with my friend, and she advised me to communicate openly with my child. Keeping the truth hidden from the child is not a sustainable solution, especially considering that the child is already sensing something. We should be honest with the child, explaining the situation clearly, and assuring them that both parents will continue to love them just as before, providing them a sense of security.
    \\
    \hline
    \textbf{Response generation} 
    \par \textit{Reference}: Hmm, your friend gave you some great advice. Trust that through timely communication, your child will be able to understand your situation. Do you have any other questions? Do you have any family members with mental health conditions?
    \par \textit{D4}: \textbf{[Empathy]}Hmm, do you have any family members who have mental health disorders?
    \par \textit{SEO}: \textbf{[genetic, affirmation, provide suggestions]}Hmm, it seems like your family and friends are very supportive of you, and engaging in conversations with them is indeed a good solution. Now, regarding your question, do you have any family members who have mental health disorders?
    \par \textit{SEO (golden encoder)}: \textbf{[genetic, affirmation, provide suggestions]$^*$} Hmm, I can tell that you care deeply about your child, and effective communication is indeed crucial. It's important to engage in conversations with your family and friends as they will likely provide you with support. Now, regarding your question, do you have any family members who have mental health disorders?
    \\
    \hline
    \end{tabular}
    \caption{Case Analysis. The predicted labels, generated by  \textit{D4} and \textit{SEO} and labels provided to the generation model before generation for \textit{SEO (golden encoder)} are enclosed within square brackets. } 
    \label{tab:case_anapysis}
\end{table*}

\section{Ontology List}

Presented in \Cref{tab:ontology} is a comprehensive exposition of the overarching construct denoted as \textbf{SEO}, which consists of two aspects: \textbf{S}ymptom-related for TOD and \textbf{E}mpathy-related for chit-chat \textbf{O}ntology. The facet of symptom-related ontology delineates ten core symptoms, wherein each of these core symptoms is associated with a spectrum of 1 to 6 relevant fine-grained symptoms. On the other hand, the empathy-related ontology segment encompasses four discerning empathetic strategies, each of which is compounded by 1 to 3 interrelated skills. This multifaceted ontology construct serves as a cornerstone, furnishing an elaborate framework to generate reliable and empathetic responses in a conversational context. 

\begin{table*}[htbp]
\resizebox{1\linewidth}{!}{
\begin{tabular}{|l|l|l|l|l|}
\hline
Aspects & Core symptoms & Fine-grained symptoms & Remark & Occurrence \\ \hline
\multicolumn{1}{|l|}{\multirow{38}{*}{TOD}} & Cause & cause & The primary reason for patients seeking assistance. & 1245 \\ 
\cline{2-5} 
& \multirow{3}{*}{Mood} & whether & Whether patients in a bad mood & 979 \\
\cline{3-5} 
& & duration & & 523 \\
\cline{3-5} 
& & morning depression & Feel more sad in the morning or at night & 359 \\ 
\cline{2-5} 
& \multirow{4}{*}{Interest} & whether & Does the patient has low interest & 175 \\
\cline{3-5} 
& & duration & & 223 \\ 
\cline{3-5} 
& & range & Past hobbies or all things. & 1178 \\
\cline{3-5} 
& & indifferent & Lack of emotional experience. & 320 \\
\cline{2-5} 
& \multirow{4}{*}{Social Function} & life affair & The function of dealing with life affairs & 397 \\
\cline{3-5} 
& & study \& work & & 507\\
\cline{3-5} 
& & social contact & Whether to contact/talk to family and friends. & 765 \\
\cline{3-5} 
& & social interact & Whether patients deliberately avoid social interaction. & 253 \\
\cline{2-5} 
& \multirow{5}{*}{Mental Status} & decreased concentration & & 267 \\
\cline{3-5}
& & memory loss & & 307 \\
\cline{3-5} 
& & tired & & 987 \\
\cline{3-5} 
& & difficulty in decision & & 323 \\
\cline{3-5} 
& & decline in self-confidence & & 852 \\
\cline{2-5} 
& \multirow{6}{*}{Sleep} & whether & Does the patient has sleep problems & 1288 \\
\cline{3-5} 
& & difficulty in falling sleep & & 469 \\
\cline{3-5} 
& & light sleep & & 413 \\
\cline{3-5} 
& & wake up early & & 331 \\
\cline{3-5} 
& & sleep too short & & 284 \\
\cline{3-5} 
& & dream & & 350 \\
\cline{2-5} 
& \multirow{4}{*}{Appetite} & whether & Does the patient has appetite problems & 1258 \\
\cline{3-5} 
& & loss of appetite & & 76 \\
\cline{3-5} se
& & overeating & & 135\\
\cline{3-5} 
& & significant weight change & & 755\\
\cline{2-5} 
& \multirow{3}{*}{Somatic Symptoms} & pyschomotor agitation & Excessive excitement and loquacity. & 547 \\
\cline{3-5} 
& & psychomotor retardation & Slow response & 524 \\
\cline{3-5} 
& & physical discomfort & & 1218 \\
\cline{2-5} 
& \multirow{6}{*}{Suicide} & self-harm-tendency & & 549 \\
\cline{3-5} 
& & suicidal tendency & & 492 \\
\cline{3-5} 
& & suicidal behavior & & 305\\
\cline{3-5} 
& & hopelessness & & 574 \\
\cline{3-5} 
& & guilt & & 498 \\
\cline{3-5} 
& & low self-worth & & 506 \\
\cline{2-5} 
& \multirow{2}{*}{Screening} & mania & Is it irritable and prone to disputes  & 732\\
\cline{3-5} 
& & genetic & & 447\\
\hline
\multicolumn{1}{|l|}{\multirow{7}{*}{Chitchat}} & Starting/ending & starting/ending & & 4302 \\
\cline{2-5} 
& \multirow{2}{*}{Give information} & question & Inquire about personal information of patients & 2309 \\
\cline{3-5} 
& & restatement & & 1776 \\
\cline{2-5} 
& \multirow{3}{*}{Show empathy} & reflection & \begin{tabular}[c]{@{}l@{}}The doctor demonstrated an understanding of the \\ emotions experienced by the patient.\end{tabular}  & 1145 \\
\cline{3-5} 
& & self-disclosure & The doctor expressed their own emotions and viewpoints. & 55 \\ \cline{3-5} 
& & affirmation & The doctor positively acknowledged and recognized the patient. & 1973 \\ \cline{2-5} 
& Seek help & provide suggestions & & 2597 \\ \hline
\end{tabular}}
\caption{The Definition of SEO}
\label{tab:ontology}
\end{table*}

\end{document}